\documentclass[runningheads]{llncs}
\usepackage{cite}
\usepackage{amsmath,amssymb,amsfonts}
\usepackage{graphicx}
\usepackage{textcomp}
\usepackage{xcolor}
\usepackage[linesnumbered,ruled,vlined,boxed]{algorithm2e}
\usepackage[justification=centering]{caption}
\usepackage{multirow}

\def\BibTeX{{\rm B\kern-.05em{\sc i\kern-.025em b}\kern-.08em
    T\kern-.1667em\lower.7ex\hbox{E}\kern-.125emX}}
\begin{document}

\title{Quantum-based Feature Selection for Multi-Classification Problem in Complex Systems with Edge Computing}
\titlerunning{}
\author{{Wenjie Liu}\inst{1}\and Junxiu Chen\inst{2}\and Yuxiang Wang\inst{3} \and Peipei Gao \inst{4}\and Zhibin Lei\inst{5}\and Xu Ma\inst{6}}
\institute{Jiangsu Engineering Center of Network Monitoring, Nanjing University of Information Science and Technology, Nanjing 210044, P.R.China\inst{1}\\
School of Computer and Software, Nanjing University of Information Science and Technology, Nanjing 210044, P.R.China\inst{2}\\
Jiangsu Engineering Center of Network Monitoring, Nanjing University of Information Science and Technology, Nanjing 210044, P.R.China\inst{3}\\
Jiangsu Engineering Center of Network Monitoring, Nanjing University of Information Science and Technology, Nanjing 210044, P.R.China\inst{4}\\
Hong Kong Applied Science and Technology Research Institute(ASTRI),Hong Kong 999077, China\inst{5}\\
School of Software, Qufu Normal University,Shandong 273165, China\inst{6}}

\maketitle

\begin{abstract}
The complex systems with edge computing require a huge amount of multi-feature data to extract appropriate insights for their decision-making, so it is important to find a feasible feature selection method to improve the computational efficiency and save the resource consumption. In this paper, a quantum-based feature selection algorithm for the multi-classification problem, namely QReliefF, is proposed, which can effectively reduce the complexity of algorithm and improve its computational efficiency. First, all features of each sample are encoded into a quantum state by performing operations $CMP$ and $R_y$, then the amplitude estimation is applied to calculate the similarity between any two quantum states (i.e., two samples). According to the similarities, the Grover-Long method is utilized to find the nearest $k$ neighbor samples, and then the weight vector is updated. After a certain number of iterations through the above process, the desired features can be selected with regards to the final weight vector and the threshold $\tau$. Compared with the classical ReliefF algorithm, our algorithm reduces the complexity of similarity calculation from $O(MN)$ to $O(M)$, the
complexity of finding the nearest neighbor from $O(M)$ to $O(\sqrt M)$, and resource consumption from $O(MN)$ to $O(MlogN)$. Meanwhile, compared with the quantum Relief algorithm, our algorithm is superior in finding the nearest neighbor, reducing the complexity from $O(M)$ to $O(\sqrt M)$. Finally, in order to verify the feasibility of our algorithm, a simulation experiment based on Rigetti with a simple example is performed.
\end{abstract}


\section{Introduction}
Complex systems~\cite{Alfieri2013Complex} are nonlinear systems composed of agents that can act with local environmental information, which require big data to extract appropriate insights for their decision-making.
In the cloud computing~\cite{Armbrust2010A}\cite{Chen2016Efficient}\cite{Xu2018AnIoTOriented}\cite{Qi2019AQoS}, the data transmission delay between the data sources and the cloud centers is problematic for many complex systems where responses are usually required to be time critical or real-time. Instead, a recently emerging computation paradigm, edge computing~\cite{Mao2017A}\cite{Xu2019AnEnergy}\cite{Qi2018ATwo}\cite{Xu2019BeCome}, is promising to cater for these requirements, as edge computing resources are deployed data sources which support time critical or real-time data processing and analysis. As we all know, the computing resources and storage resources of most intelligent terminals are very limited, which places higher requirements on the computing performance of algorithms, especially machine learning algorithms, in complex systems with edge computing.

Machine learning~\cite{Mohri12}\cite{Xu2019ABlockchain} continuously improves its performance through ``experience'', where experience generally originates from massive data. At present, many machine learning algorithms based on massive data~\cite{Manogaran2018}\cite{Qi2019Time}\cite{Xu2019Acomputation} have been proposed. In the practical scenario, the amount of data available for training is getting larger and larger, while the characteristics of data are becoming more and more abundant. Those data with redundant or unrelated features will cause the problem of ``curse of dimensionality''~\cite{Howard1966}, which greatly increases the computational complexity of the algorithm. One of the possible solutions is the dimension reduction~\cite{Roweis2000Nonlinear}, and the other is the feature selection~\cite{Liu2012Feature}.

Relief algorithm~\cite{Kira92} is a well-known feature selection algorithm for the two-classification problem. It is widely used because of its excellent classification effect. However, the limitation of this algorithm is that it can only perform the binary classification, and the efficiency of the algorithm will be greatly affected when the data size and feature size increase. To extend the application of the algorithm, Kononenko \cite{Kononenko94} proposed a new feature selection algorithm for the multi-classification problem, namely ReliefF algorithm. It has the advantages of simple principle, convenient implementation and good results, and has been widely applied in various fields \cite{Zhang08}\cite{Kong12}\cite{Moore15}.

On the other hand, since Benioff \cite{Benioff80} and Feynman \cite{Feynman82} explored the theoretical possibilities of quantum computing, some excellent results have been proposed one after another. For instance, Shor's algorithm \cite{Shor99} solves the problem of integer factorization in polynomial time. Grover's algorithm \cite{Grover97} has a quadratic speedup to the problem of conducting a search through some unstructured database. These excellent results have prompted people to think about how to apply this computing power into machine learning algorithms. And thus a new research hotspot, quantum machine learning~\cite{Qu2018Novel}\cite{Liu2019Unitary}\cite{Liu2019QuantumBased}\cite{Lamata2019Quantum}\cite{Qu2019QuantumImage}\cite{Qu2019Matrix}, has gradually formed. Although quantum technology provides a certain improvement in storage and computing power, the ``curse of dimensionality'' problem still exists in quantum machine learning. Therefore, the quantum-based dimensionality reduction method still has important research value. In 2018, Liu et al.\cite{Liu18} proposed a quantum Relief algorithm (namely QRelief algorithm) for the two-classification problem, which reduces the complexity of similarity calculation from $O(MN)$ to $O(M)$.

As we know, in the application scenario of edge computing, there are various multi-classification problems based on distributed, massive and large-feature data. The objective of this study is to design a feasible feature selection method which can effectively get rid of redundant or unrelated features in machine learning, reducing the computation load of intelligent terminals, and thus meet the requirement of real-time data
processing and analysis in edge computing. In this paper, we introduce some quantum technologies (such as \emph{CMP} operation, amplitude estimation and Grover-Long method), and propose a quantum-based feature selection algorithm, namely QReliefF algorithm, for the multi-classification problem.

The main contributions of our work are:

(1) A quantum method is proposed to solve the problem of feature selection for the multi-classification problem in complex systems
with edge computing. The proposed method fully demonstrates the quantum parallel processing capabilities that classical computing cannot match, and significantly reduces the computational complexity of the algorithm.

(2) The problem of finding nearest neighbor samples is firstly transformed into the similarity calculation of two quantum states (i.e., calculating their inner product) in quantum computing, and the Grover-long method is utilized to speed up the search of the targets.

(3) A simulation experiment based on Rigetti is performed to verify the feasibility of our algorithm.

The outline of this paper is as follows: The classic ReliefF algorithm is briefly reviewed in Sect. 2, and the proposed quantum ReliefF algorithm is proposed in detail in Sect. 3. Then, we illustrate the process of the algorithm with a simple example in Sect. 4, and perform the simulation experiment on Rigetti in Sect. 5. Subsequently, the efficiency of the algorithm is analyzed in Sect. 6, and the brief conclusion and discussion are summarized in the last section.

\section{Review of ReliefF algorithm}
ReliefF algorithm~\cite{Kononenko94} is a feature selection algorithm which is used to handle the multi-classification problem. Before introducing our proposed quantum algorithm, let us review the detailed process of the algorithm.

Without loss of generality, suppose there are $M$ samples with $N$ features, and they can be divided into $P$ classes:
\begin{equation}
{C_p} = \left\{ {{v_q}\left| {{v_q} \in {\mathbb{R}^N},q = 1,2, \ldots ,{M_p}} \right.} \right\}, p \in \left\{ {1,2, \ldots ,P} \right\},
\label{eq16}
\end{equation}
where ${v_q}$ is the $q$-th $N$-feature sample that belongs to Class ${C_p}$, ${v_q} = {\left( {{v_{q1}},{v_{q2}}, \ldots ,{v_{qN}}} \right)^T}$. And the weight vector of $N$ features $WT = {(w{t_1},w{t_2}, \cdots ,w{t_N})^T}$ is initialized to all zeros, the upper limit of iteration is $T$, and the relevance threshold (that differentiate the relevant and irrelevant features) is $\tau$ (${\rm{0}} \le \tau  \le 1$). The main steps of ReliefF algorithm are as follows (its pseudo code can be seen in Algorithm~\ref{alg1}).

\renewcommand{\algorithmcfname}{Algorithm}
\begin{algorithm}
\SetAlgoNoLine
\caption{ReliefF algorithm}
\label{alg1}
\emph{Init $WT = {(0, \cdots ,0)^T}$}

\For{$t = 1$ \KwTo $T$}{

\emph{Pick a sample $u$ randomly }

\emph{Find $k$ nearest neighbor samples ${H_j}$ from the same class of sample $u$.}

\For{${C_p} \ne class(u)$}{
\emph{Find $k$ nearest neighbor samples ${M_j}(C_p)$ from the different Class $C_p$ }
}
\For{$i = 1$ \KwTo $N$}{
\emph{$WT\left[ i \right] = WT\left[ i \right] - \sum\limits_{j = 1}^k {diff(i,u,{H_j})} {\kern 1pt} {\kern 1pt} {\kern 1pt}  + \sum\limits_{{C_p} \notin class(u)} {[{{p(C_p)} \over {1 - p(class(u))}}\sum\limits_{j = 1}^k {diff(i,u,{M_j}(C_p))} ]} $}
}}
\emph{Select the most relevant features according to $WT$ and $\tau $}
\end{algorithm}

At each iteration, ReliefF randomly selects a sample $u$, and then searches for $k$ nearest neighbor samples by cosine distance from each class. The closest same-class sample is called ${H_j}$, and the closest different-class sample is called ${M_j}(C_p)$, where $j = \{ 1,2, \cdots ,k\} $. The updating weight vector formula is shown as follows,
\begin{equation}
\begin{gathered}
  WT\left[ i \right] = WT\left[ i \right] - \sum\limits_{j = 1}^k {diff(i,u,{H_j})}  \hfill \\
   + \sum\limits_{{C_p} \notin class(u)} {[\frac{{p(C_p)}}{{1 - p(class(u))}}\sum\limits_{j = 1}^k {diff(i,u,{M_j}(C_p))} ]}  \hfill \\
\end{gathered} ,
\label{eq1}
\end{equation}
{where $p(C_p)$ represents the probability of randomly extracting samples of Class $C_p$, and the definition of $diff(i,u,v)$ function is as follows,}
\begin{equation}
diff(i,u,v) = \left\{
             \begin{array}{lcl}
             {{{\left| {u[i] - v[i]} \right|} \over {\max (i) - \min (i)}}}{\kern 4pt} i{\kern 2pt}is{\kern 2pt}continuous\\
             0{\kern 28pt}  {u_i} = {v_i} \\
             1{\kern 30pt} {u_i} \ne {v_i}
             \end{array}
        .\right.
\label{eq2}
\end{equation}
After iterating $T$ times, the final weight vector is obtained. Through the relevance threshold $\tau$, we can retain relevant features and discard irrelevant features.

ReliefF algorithm is an extension of Relief algorithm that extends the two-classification problem to multi-classification scenario. However, with the increase of category size, sample size and sample features, ReliefF algorithm will also face with the problem of ``dimension disaster'', and the speed of the algorithm will also drop sharply. So, how to improve the efficiency of ReliefF algorithm becomes an urgent problem to be solved.

\section{The proposed QReliefF algorithm}
In order to implement the feature selection for the multi-classification problem in complex systems with edge computing, a feasible quantum ReliefF algorithm is introduced in this section. Suppose the sample sets ${{ C}_p} = \left\{ {
{v_q} = {\left( {{v_{q1}},{v_{q2}}, \ldots ,{v_{qN}}} \right)^T}
\left| {{{ v}_q} \in {\mathbb{R}^N},q = 1,2, \ldots ,{M_p}} \right.} \right\}$ ($p$ represents the category of classification, $p \in \left\{ {1,2, \ldots ,P} \right\}$), the weight vector $WT$, the upper limit $T$, and  the relevance threshold $\tau$ are the same as classical ReliefF algorithm defined in Sect. II. Different from the classical one, all the features of each sample are represented as a quantum superposition state, thus the problem of finding nearest neighbor samples is transformed into the similarity calculation of two quantum states (i.e., calculating their inner product). And the similarity between any two samples can be calculated in parallel in the way of quantum mechanics. Algorithm~\ref{alg2} describes the process of our algorithm in detail, and the specific steps are as bellow.

\renewcommand{\algorithmcfname}{Algorithm}
\begin{algorithm}
\SetAlgoNoLine
\caption{Quantum ReliefF algorithm}
\label{alg2}
\emph{Init $WT = {(0, \cdots ,0)^T}$}

\emph{Normalized the sample sets: ${C_p} \to {{\bar C}_p}$}

\emph{prepare quantum states for all samples by operations $CMP$ and $R_y$, respectively.
${\left| {{\phi _p}} \right\rangle _q} = {1 \over {\sqrt N }}\left| q \right\rangle \sum\limits_{i = 0}^{N - 1} {\left| i \right\rangle \left| 1 \right\rangle \left( {\sqrt {1 - {{\left| {{\bar v_{q{\kern 1pt} i}}} \right|}^2}} \left| 0 \right\rangle  + {\bar v_{q{\kern 1pt} i}}\left| 1 \right\rangle } \right)} $}

\For{$t = 1$ \KwTo $T$}{

\emph{Select a state $\left| \phi  \right\rangle $ from $\left\{ {{{\left| {{\phi _p}} \right\rangle }_q}} \right\}$} randomly which corresponds to $u$

\emph{Perform swap operation on $\left| \phi  \right\rangle $ and obtain
$\left| \varphi  \right\rangle  = {1 \over {\sqrt N }}\left| l \right\rangle \sum\limits_{i = 0}^{N - 1} {\left| i \right\rangle \left( {\sqrt {1 - {{\left| {{\bar u_i}} \right|}^2}} \left| 0 \right\rangle  + {\bar u_i}\left| 1 \right\rangle } \right)} \left| 1 \right\rangle $}

\emph{The similarity information coded into quantum state ${\left| \beta  \right\rangle _p} = {1 \over {\sqrt {{M_p}} }}\sum\limits_{q = 1}^{{M_p}} {\left| q \right\rangle } \left| {{\bar v_q} - \bar u} \right\rangle $} through $swap{\kern 1pt} {\kern 1pt} {\kern 1pt} {\kern 1pt} test$, the inner product and amplitude estimation operations

\emph{The nearest $k$ samples in each class are obtained by Grover-Long method}

\For{$i = 1$ \KwTo $N$}{
\emph{$WT\left[ i \right] = WT\left[ i \right] - \sum\limits_{j = 1}^k {diff(i,\bar u,{H_j})}  + \sum\limits_{{\bar C_p} \notin class(\bar u)} {[{{p(\bar C_p)} \over {1 - p(class(\bar u))}}\sum\limits_{j = 1}^k {diff(i,\bar u,{M_j}(\bar C_p))} ]} $}
}
}
\emph{$\overline {WT}  = \left( {{1 \mathord{\left/
 {\vphantom {1 T}} \right.
 \kern-\nulldelimiterspace} T}} \right)WT$}

\For{$i = 1$ \KwTo $N$}{

\eIf{$\left( {\overline {W{T_i}}  \ge \tau } \right)$}{
\emph{The i-th feature is relevant}\;
}{
\emph{The i-th feature is not relevant}\;
}}
\end{algorithm}

\subsection{State preparation}
In order to store classical information in quantum states, we need to normalize the sample sets:
\begin{equation}
{C_p} \to {{\bar C}_p}= \left\{ {
{\bar v_q} = {\left( {{\bar v_{q1}},{\bar v_{q2}}, \ldots ,{\bar v_{qN}}} \right)^T}} \right\},
\label{eq34}
\end{equation}
where
\begin{equation}
{\bar v_{qi}} = {{{v_{qi}}} \over {\sqrt {\sum\limits_{i = 1}^N {{{\left| {{v_{qi}}} \right|}^2}} } }}, i = 1,2, \ldots ,N.
\label{eq33}
\end{equation}
Obviously, ${{\bar v}_{qi}}$ is a real number, and ${{\bar v}_{qi}} \in \left( {0,1} \right)$.
Then we prepare the initial quantum states as below,
\begin{equation}
{\left| {{\phi _p}} \right\rangle _q} = \frac{1}{{\sqrt N }}\left| q \right\rangle \sum\limits_{i = 0}^{N - 1} {\left| i \right\rangle \left| 1 \right\rangle \left( {\sqrt {1 - {{\left| {{\bar v_{q{\kern 1pt} i}}} \right|}^2}} \left| 0 \right\rangle  + {\bar v_{q{\kern 1pt} i}}\left| 1 \right\rangle } \right)} ,
\label{eq3}
\end{equation}
where ${\left| {{\phi _p}} \right\rangle _q}$ corresponds to the quantum state of the $q$-th sample that belongs to Class ${\bar C_p}$, and ${\bar v_{q{\kern 1pt} i}}$ represents the $i$-th eigenvalue of the $q$-th sample. Assume our initial state is $\left| q \right\rangle {\left| 0 \right\rangle ^{ \otimes n}}\left| 1 \right\rangle \left| 0 \right\rangle (n = \left\lceil {{{\log }_2}(N)} \right\rceil )$, the construction scheme of the quantum state ${\left| {{\phi _p}} \right\rangle _q}$ includes the following steps.

\begin{figure}[htbp]
\centerline{\includegraphics[height=2.5cm,width=8cm,]{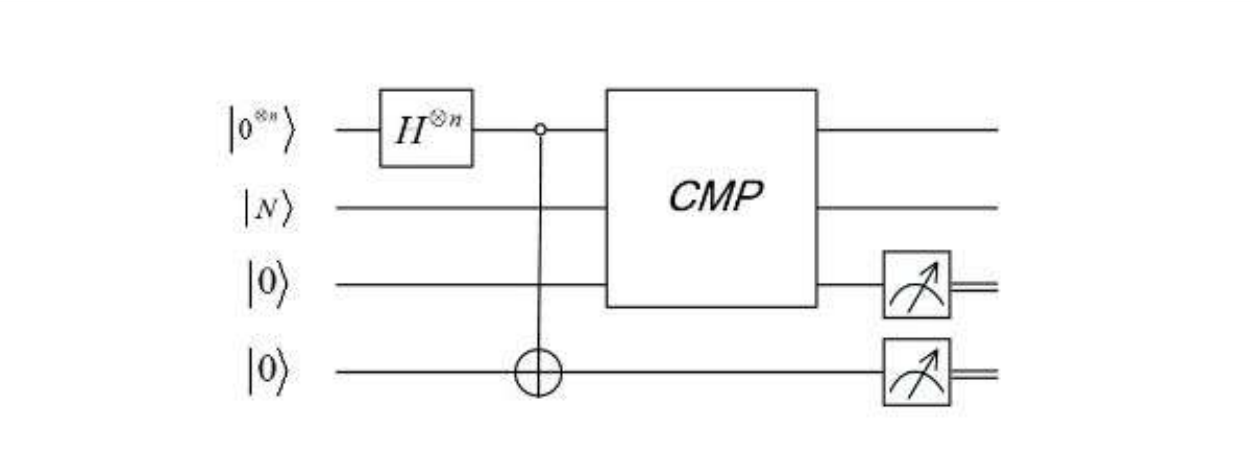}}
\caption{Quantum circuit of getting ${{\rm{1}} \over {\sqrt N }}\sum\limits_{i = 0}^{N-1} {\left| i \right\rangle }$, and \emph{H} is the Hadamard operation, $\circ $ represents the control qubit conditional being set to zero.}
\label{fig01}
\end{figure}

\begin{figure}[htbp]
\centerline{\includegraphics[height=2.5cm,width=8cm,]{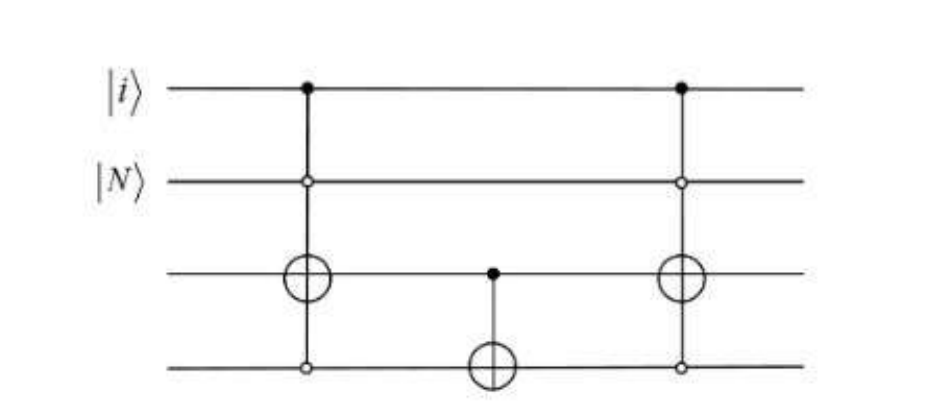}}
\caption{Quantum circuit of \emph{CMP} operation, and $\bullet$ represents the control qubit conditional being set to one.}
\label{fig02}
\end{figure}

First, we perform Hadamard and \emph{CMP} operations for ${\left| 0 \right\rangle ^{ \otimes n}}$ and get a new state:
\begin{equation}
{\left| 0 \right\rangle ^{ \otimes n}} \xrightarrow{\emph{H} {\kern 2pt} and {\kern 2pt}\emph{CMP}{\kern 2pt} operations}{{\rm{1}} \over {\sqrt N }}\sum\limits_{i = 0}^{N-1} {\left| i \right\rangle },
\label{eq4}
\end{equation}
and its circuit diagram is shown in Fig.~\ref{fig01}, where the definition of \emph{CMP} operation is
\begin{equation}
CMP\left| i \right\rangle \left| N \right\rangle \left| 0 \right\rangle  = \left\{ {
\begin{array}{*{20}{c}}
{\left| i \right\rangle \left| N \right\rangle \left| 0 \right\rangle ,i < N}\\
{\left| i \right\rangle \left| N \right\rangle \left| 1 \right\rangle ,i  \ge  N}
\end{array}} .\right.
\label{eq5}
\end{equation}
The function of \emph{CMP} operation is to cut the quantum state larger than $N$, and its circuit diagram is shown in Fig.~\ref{fig02}. $\left| i \right\rangle $ and $\left| N \right\rangle $ represent a single qubit. The implementation of \emph{CMP} operation needs to repeatedly implement such a circuit $n$ times. After measurement, if the lowest register is $\left| 1 \right\rangle $, it means that $i > N$.

Next, we perform the unitary rotation operation $R_y$:
\begin{equation}
{R_y}\left( {2{{\sin }^{ - 1}}{\bar v_{qi}}} \right) = \left[ {
\begin{array}{*{20}{c}}
  {\sqrt {1 - {{\left| {{\bar v_{qi}}} \right|}^2}} }&{ - {\bar v_{qi}}} \\
  {{\bar v_{qi}}}&{\sqrt {1 - {{\left| {{\bar v_{qi}}} \right|}^2}} }
\end{array}} \right],
\label{eq6}
\end{equation}
on the last qubit to obtain our target quantum state ${\left| {{\phi _p}} \right\rangle _q}$,
\begin{equation}
\frac{1}{{\sqrt N }}\left| q \right\rangle \sum\limits_{i = 0}^{N - 1} {\left| i \right\rangle \left| 1 \right\rangle \left( {\sqrt {1 - {{\left| {{\bar v_{q{\kern 1pt} i}}} \right|}^2}} \left| 0 \right\rangle  + {\bar v_{q{\kern 1pt} i}}\left| 1 \right\rangle } \right)} .
\label{eq7}
\end{equation}

\subsection{Similarity calculation}
After the state preparation, the information of the samples is encoded into the quantum superposition state $\left\{ {{{\left| {{\phi _p}} \right\rangle }_q}} \right\}$. In this paper, we use the cosine distance to define the similarity between the random sample $\bar u$ and other sample (e.g., $\bar v_q$),
\begin{equation}
s\left( {\bar u,{\bar v_q}} \right) \buildrel\textstyle.\over= {\left| {\cos \theta } \right|^2} = {{{{\left| {\left\langle {\bar u}
 \mathrel{\left | {\vphantom {\bar u {{\bar v_q}}}}
 \right. \kern-\nulldelimiterspace}
 {{{\bar v_q}}} \right\rangle } \right|}^2}} \over {{{\left| \bar u \right|}^2} \cdot {{\left| {{\bar v_q}} \right|}^2}}}.
\label{eq31}
\end{equation}
Referring to Eqs.~(\ref{eq34}) and~(\ref{eq33}), ${\left| {\bar u} \right|^2}$ and ${\left| {{\bar v_q}} \right|^2}$ are 1, Eq.~\ref{eq31} can be simplified as follows,
\begin{equation}
s(\bar u,{\bar v_q}) = {\left| {\left\langle {\bar u}
 \mathrel{\left | {\vphantom {\bar u {{\bar v_q}}}}
 \right. \kern-\nulldelimiterspace}
 {{{\bar v_q}}} \right\rangle } \right|^2}.
\label{eq32}
\end{equation}

First, $\left| \phi  \right\rangle $ (i.e., the sample $\bar u$) is randomly selected from $\left\{ {{{\left| {{\phi _p}} \right\rangle }_q}} \right\}$ which is the $l$-th sample in Class ${\bar C_p}$, as shown in the following equation,
\begin{equation}
\left| \phi  \right\rangle  = \frac{1}{{\sqrt N }}\left| l \right\rangle \sum\limits_{i = 0}^{N - 1} {\left| i \right\rangle \left| 1 \right\rangle \left( {\sqrt {1 - {{\left| {{\bar u_i}} \right|}^2}} \left| 0 \right\rangle  + {\bar u_i}\left| 1 \right\rangle } \right)} .
\label{eq8}
\end{equation}
Then, a $swap$ operation is performed on $\left| \phi  \right\rangle $ to get
\begin{equation}
\left| \varphi  \right\rangle  = \frac{1}{{\sqrt N }}\left| l \right\rangle \sum\limits_{i = 0}^{N-1} {\left| {i} \right\rangle \left( {\sqrt {1 - {{\left| {{\bar u_i}} \right|}^2}} \left| 0 \right\rangle  + {\bar u_i}\left| 1 \right\rangle } \right)\left| 1 \right\rangle } .
\label{eq9}
\end{equation}
Next, a \emph{swap test} (its circuit is given in Fig.\ref{fig03}) is performed on $\left( {\left| \varphi  \right\rangle ,{\kern 1pt} {\kern 1pt} {\kern 1pt} {\kern 1pt} {\kern 1pt} {\kern 1pt} {{\left| {{\phi _p}} \right\rangle }_q}} \right)$ and obtain
\begin{equation}
\left| \psi  \right\rangle = {1 \over 2}\left| 0 \right\rangle (\left| \varphi  \right\rangle {\left| {{\phi _p}} \right\rangle _q} + {\left| {{\phi _p}} \right\rangle _q}\left| \varphi  \right\rangle ) + {1 \over 2}\left| 1 \right\rangle (\left| \varphi  \right\rangle {\left| {{\phi _p}} \right\rangle _q} - {\left| {{\phi _p}} \right\rangle _q}\left| \varphi  \right\rangle ).
\label{eq10}
\end{equation}

\begin{figure}[htbp]
\caption{Quantum circuit of \emph{swap test} operation, and the symbol of two crosses connected by a line represents the \emph{swap} operation.}
\label{fig03}
\end{figure}

From Eq.~(\ref{eq10}), we know the probability of measurement result being $\left| 1 \right\rangle $ is
\begin{equation}
\begin{gathered}
P_q^l(1) = \left\langle \psi  \right| \left| 1 \right\rangle \langle 1| \otimes I \otimes I \left| \psi  \right\rangle   \hfill \\
\;\;\;\;\;\;\;\;\;= \left[ {{1 \over 2}\left\langle 1 \right|\left( {\left\langle \varphi  \right|{{\left\langle {{\phi _p}} \right|}_q} - {{\left\langle {{\phi _p}} \right|}_q}\left\langle \varphi  \right|} \right)} \right]\left| 1 \right\rangle \langle 1| \otimes I \otimes I \hfill \\
\;\;\;\;\;\;\;\;\;\;\;\;\;\left[ {{1 \over 2}\langle 1|\left( {\left| \varphi  \right\rangle {{\left| {{\phi _p}} \right\rangle }_q} - {{\left| {{\phi _p}} \right\rangle }_q}\left| \varphi  \right\rangle } \right)} \right]  \hfill \\
\;\;\;\;\;\;\;\;\;= {1 \over 2} - {1 \over 2}{\left| {{{\left\langle {\varphi \left| {{\phi _p}} \right.} \right\rangle }_q}} \right|^2},\hfill \\
\end{gathered}
\label{eq28}
\end{equation}
In addition, the inner product between $\left|\varphi  \right\rangle$ and ${{\left| {{\phi _p}} \right\rangle}_q}$ (i.e., the prepared state) can be calculated
as below,
\begin{equation}
{\left\langle {\varphi } \mathrel{\left | {\vphantom {\varphi  {{\phi _p}}}} \right. \kern-\nulldelimiterspace} {{{\phi _p}}} \right\rangle _q} = {1 \over N}\sum\limits_i {\left( {{\bar u_i}} \right) * {\bar v_{qi}}} = {1 \over N}\left\langle {\bar u} \mathrel{\left | {\vphantom {\bar u {{\bar v_q}}}} \right. \kern-\nulldelimiterspace} {{{\bar v_q}}} \right\rangle.
\label{eq29}
\end{equation}
Combining Eq.~(\ref{eq28}) with Eq.~(\ref{eq29}), we can get the similarity between samples $\bar u$ and $\bar v_q$,
\begin{equation}
s(\bar u,{\bar v_q}) = \left( {1 - 2P_q^{l\left( {{\bar C_p}} \right)}(1)} \right){N^2}
\label{eq30}
\end{equation}
Since $N$ is a constant value and $\left\langle {\bar u} \mathrel{\left | {\vphantom {\bar u {{\bar v_q}}}} \right. \kern-\nulldelimiterspace} {{{\bar v_q}}} \right\rangle $ is the angle cosine between the random sample $\bar u$ and other sample ${\bar v_q}$ (e.g., in Class ${\bar C_p}$), then the smaller $s(\bar u,{\bar v_q})$ is, the smaller cosine distance is, which indicates that these two samples are more similar.

Then we can rewrite Eq.~(\ref{eq10}) as bellow:
\begin{equation}
{\left| \alpha  \right\rangle _p} = {1 \over {\sqrt {{M_p}} }}\sum\limits_{q = 1}^{{M_p}} {\left| q \right\rangle } \left( {\sqrt {1 - s(\bar u,{\bar v_q})} \left| 0 \right\rangle  + \sqrt {s(\bar u,{\bar v_q})} \left| 1 \right\rangle } \right).
\label{eq11}
\end{equation}

\subsection{Finding the nearest neighbor samples}
\centerline{\includegraphics[height=2.5cm,width=8cm,]{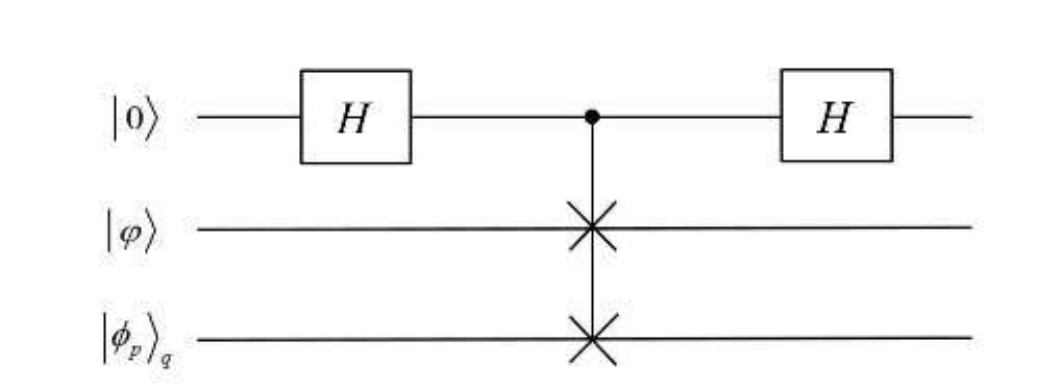}}

\begin{figure}[htbp]
\centerline{\includegraphics[height=2.5cm,width=8cm,]{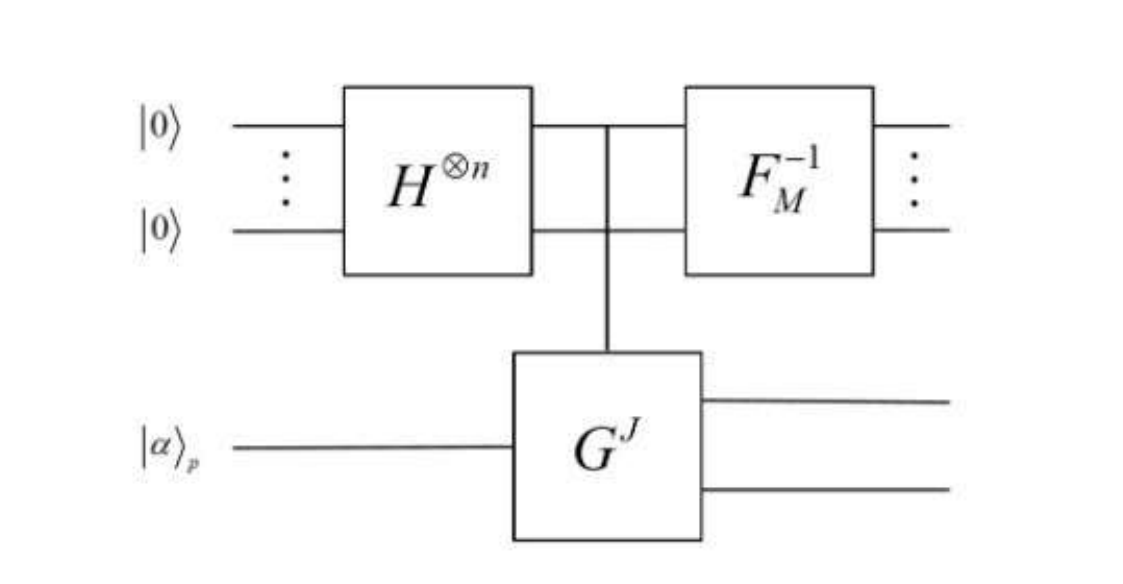}}
\caption{Quantum circuit of amplitude estimation operation, ${G^J}$ represents $J$ iterations of Grover-Long method~\cite{Long2001Grover} and $F_M^{ - 1}$ represents the inverse Fourier transform~\cite{Zhou2017Quantum}.}
\label{fig3}
\end{figure}

First, the quantum amplitude estimation method~\cite{Nielsen2000quantum} is applied to store the similarity of the sample $\bar u$ and $\bar v_q$ in the last qubit,
\begin{equation}
{\left| \beta  \right\rangle _p} = {1 \over {\sqrt {{M_p}} }}\sum\limits_{q = 1}^{{M_p}} {\left| q \right\rangle } \left| {s\left( {\bar u,{{\bar v}_q}} \right)} \right\rangle ,
\label{eq12}
\end{equation}
where $p \in \left\{ {1,2, \ldots ,P} \right\}$, and its quantum circuit diagram is given in Fig.~\ref{fig3}.

\begin{figure}
\centering
\includegraphics[width=3.5in]{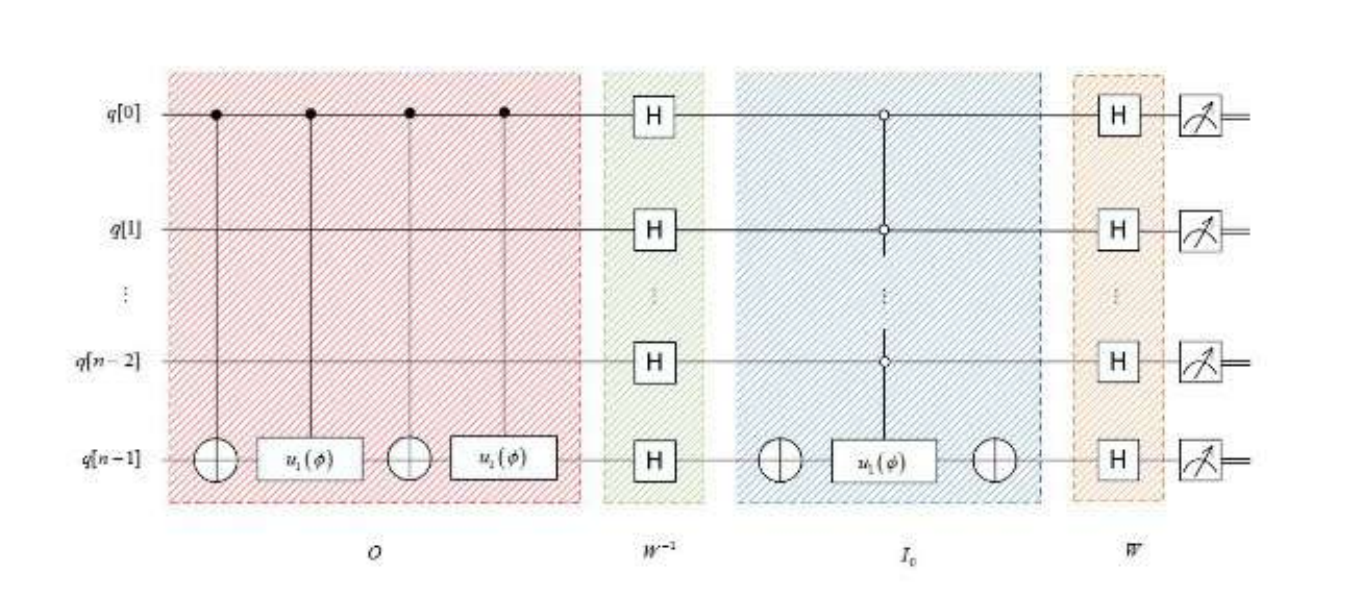}
\caption{Quantum circuit with one iteration in Grover-Long method~\cite{Long2001Grover}, and $q\left[ 0 \right]$ denotes the highest qubit, $q\left[ n-1 \right]$ denotes the lowest qubit.}
\label{fig8}
\end{figure}

In the Grover-Long method~\cite{Long2001Grover}, one iteration can be divided into four operations, i.e., $G=-W I_{0} W^{-1} O$, and its quantum circuit is shown in Fig.~\ref{fig8}. $O$ is an oracle operation which performs a phase inversion on the targets:
\begin{equation}
O = {e^{i\phi }}\left| v \right\rangle \left\langle v \right| + \sum\limits_{\tau  = 0,\tau  \ne v}^{{2^n} - 1} {\left| \tau  \right\rangle \left\langle \tau  \right|},
\end{equation}
where $ v$ is the position of ${e^{i\phi }}$ in the diagonal matrix.
The position $v $ of ${e^{i\phi }}$ is divided into two cases. If $ v $ is odd, the $u_{1}(\phi)$ operation will be applied to the lowest qubit,
\begin{equation}
u_{1}(\phi)=\operatorname{diag}\left[1, e^{i \phi}\right].
\label{eq27}
\end{equation}
If $v $ is even, $X, u_{1}(\phi), X$ operations will be applied to the lowest qubit.

Besides, $I_{0}$ is a conditional phase shift operation which performs a phase inversion on $|0\rangle$:
\begin{equation}
{I_0} = {e^{i\phi }}\left| 0 \right\rangle \left\langle 0 \right| + \sum\limits_{\tau {\rm{ = }}1}^{{2^n} - 1} {\left| \tau  \right\rangle \left\langle \tau  \right|}  = {\mathop{\rm diag}\nolimits} {\left[ {{e^{i\phi }},1, \ldots ,1} \right]_{{2^n}}},
\end{equation}
\begin{equation}
\phi=2 \arcsin \left(\frac{\sin \frac{\pi}{4 J+2}}{\sin \eta}\right).
\end{equation}
where $\sin \eta=\sqrt{\frac{M}{N}}$, $J$ represents the number of iteration.

\begin{figure}[htbp]
\centerline{\includegraphics[width=3.5in]{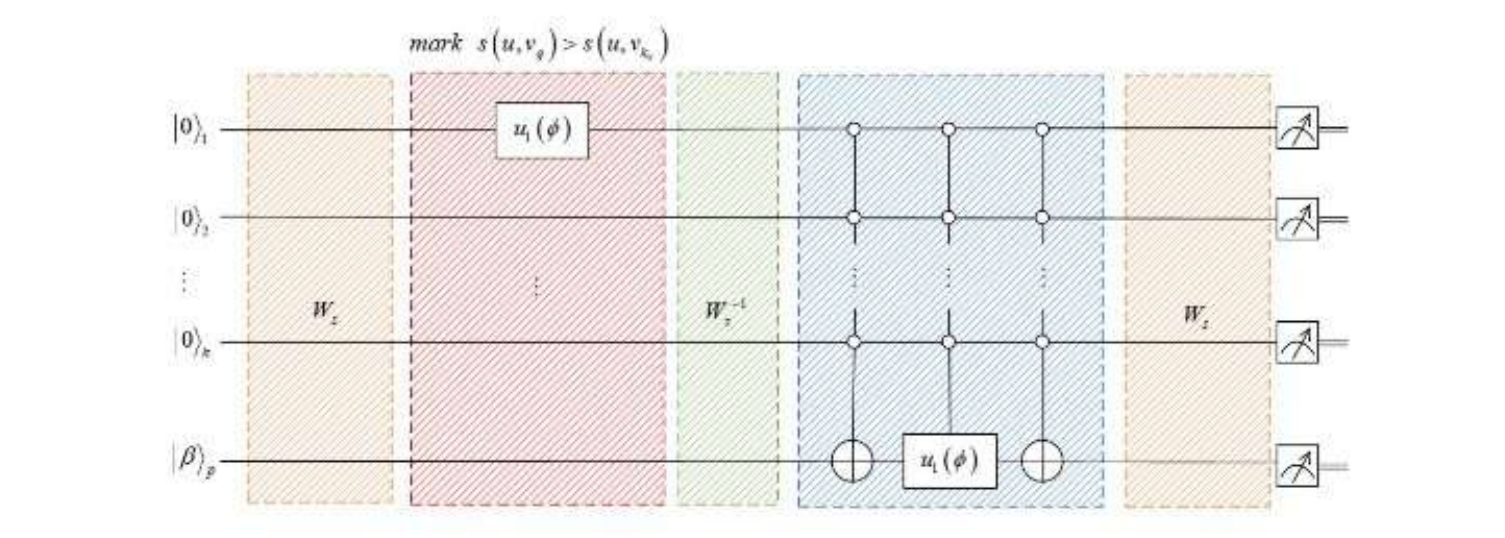}}
\caption{Quantum circuit of finding $k$ nearest neighbor samples.}
\label{fig4}
\end{figure}

Having obtained the state ${\left| \beta  \right\rangle _p}$ (see Eq.~(\ref{eq12})) through the amplitude estimation, we introduce a quantum minimum
search algorithm~\cite{Chen2019An} to find $k$ nearest neighbor samples from Class $\bar C_p$ with the time complexity of $O\left( {\sqrt {k{M_p}} } \right)$, and its quantum circuit is shown in Fig.~\ref{fig4}.

Suppose the set $K = \left\{ {{K_1},{K_2}, \cdots ,{K_k}} \right\}$ represents the $k$ nearest neighbor samples, so we should prepare $\left\lceil {\sqrt {{k}} } \right\rceil$ auxiliary qubits. As shown in Fig.~\ref{fig4}, the operator $W_s$ represents  ${W_s}{\left| \beta  \right\rangle _p}\left| {{0^{ \otimes k}}} \right\rangle  = {\left| \beta  \right\rangle _p}\left| {{K_1}} \right\rangle \left| {{K_2}} \right\rangle  \cdots \left| {{K_k}} \right\rangle $, and $u_{1}(\phi)$ is the operator defined in Eq.~(\ref{eq27}).
Let ${d_0} = s\left( {\bar u,{\bar v_1}} \right)$, we can mark ${K_x}$ when ${d_0} > s(\bar u,{\bar v_{{K_x}}}),x \in [1,k]$.
Next, we can replace $d_0$ after one iteration, where $d_0$ is $\min \{ s(\bar u,{\bar v_{{K_x}}})\} $, $x \in [1,k]$.
We repeat the above steps several times until all samples in Class $\bar C_p$ are compared. Finally, all index of $k$ nearest neighbor samples in Class $\bar C_p$ can be gotten according to the similarity.

\subsection{Updating weight vector}
After the above steps, we obtain the nearest neighbor samples (i.e., $H_j$ and ${M_j}(\bar C_p)$) of the random sample $\bar u$. Then, we update the weight vector according to the updating weight vector formula as follows,
\begin{equation}
\begin{gathered}
  WT\left[ i \right] = WT\left[ i \right] - \sum\limits_{j = 1}^k {diff(i,\bar u,{H_j})}  \hfill \\
   + \sum\limits_{{\bar C_p} \notin class(\bar u)} {[\frac{{p(\bar C_p)}}{{1 - p(class(\bar u))}}\sum\limits_{j = 1}^k {diff(i,\bar u,{M_j}(\bar C_p))} ]}  \hfill \\
\end{gathered} ,
\label{eq14}
\end{equation}
where $i \in [1,N]$.

\subsection{Feature selection}
After iterating the above steps, i.e., similarity calculation, find the nearest neighbor samples and update weight vector, $T$ times, we jump out of the algorithm's loop. Then we get a final weight vector $WT$. And the average weight vector is,
\begin{equation}
\overline {WT}  = {1 \over T}WT.
\label{eq15}
\end{equation}
Then, we make feature selection based on the final $\overline {WT} $ and threshold $\tau $. Here, $\tau $ can be chosen to retain relevant features and discard irrelevant features~\cite{Kira1992Feature}, that is to say, those features whose weight greater than $\tau $ will be selected, and those less than $\tau $ will be discarded. Here, the value of $\tau $ is determined with regards to the user's requirements and the characteristics of the problem itself (e.g., the distribution of samples, the number of features).

\section{Example}
\label{sec:4}
\begin{figure}
\centering
\includegraphics[width=2.2in]{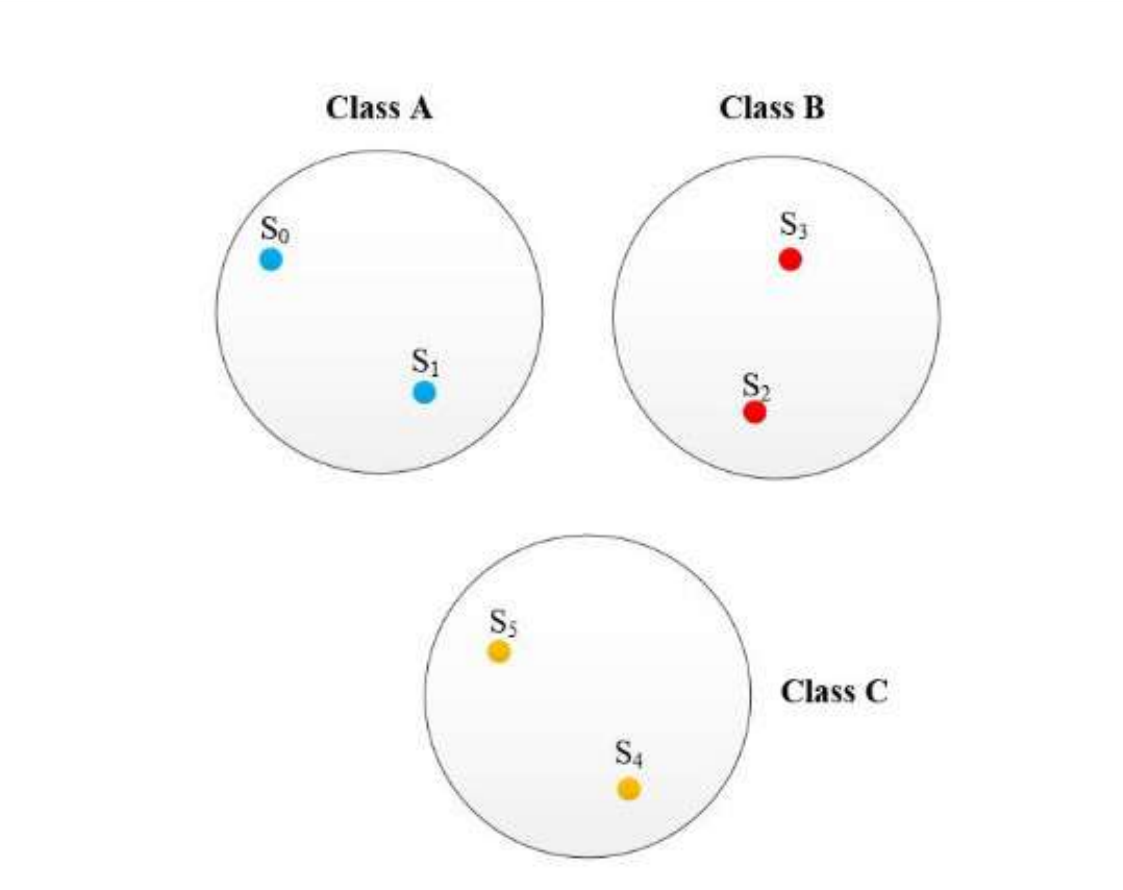}
\caption{The simple example with six samples in classes $A$, $B$ and $C$.}
\label{fig5}
\end{figure}

Suppose there are four samples(see Table~\ref{tab:1}), ${S_0}=(1,0,0,1,0,0)$, ${S_1}=(1,0,0,0,1,0)$, ${S_2}=(0,1,0,0,0,1)$, ${S_3}=(0,1,0,1,0,0)$, ${S_4}=(0,0,1,0,1,0)$, ${S_5}=(0,0,1,0,0,1)$, thus the $n$ is 3, and they belong to two classes: $A = \{ {S_0},{S_1}\}$, $B = \{ {S_2},{S_3}\}$, $C = \{ {S_4},{S_5}\}$, which is illustrated in Fig.~\ref{fig5}.
\begin{table}
\centering
\caption{The feature values of four samples. Each row represents all the feature values of a certain sample, while each column denotes a certain feature value of all the samples}
\label{tab:1}
\begin{tabular}{cllllll}
\hline\noalign{\smallskip}
 & $F_0$ & $F_1$ & $F_2$ & $F_3$& $F_4$& $F_5$  \\
\noalign{\smallskip}\hline\noalign{\smallskip}
$S_0$ & 1 & 0 & 1 & 0& 0& 0\\
$S_1$ & 1 & 0 & 0 & 0& 1& 0\\
$S_2$ & 0 & 1 & 0 & 0& 0& 1\\
$S_3$ & 0 & 1 & 0 & 1& 0& 0\\
$S_4$ & 0 & 0 & 1 & 0& 1& 0\\
$S_5$ & 0 & 0 & 1 & 0& 0& 1\\
\noalign{\smallskip}\hline
\end{tabular}
\end{table}

First, the four initial quantum states are prepared as follows,
\begin{equation}
\label{eqn:18}
\left\{ {\begin{array}{*{20}{c}}
{\left| \psi  \right\rangle _{{{\rm{S}}_{\rm{0}}}}} = \left| {{\rm{000}}} \right\rangle {\left| {\rm{0}} \right\rangle ^{ \otimes 3}}\left| 1 \right\rangle \left| 0 \right\rangle \\
{\left| \psi  \right\rangle _{{{\rm{S}}_{\rm{1}}}}} = \left| {{\rm{001}}} \right\rangle {\left| {\rm{0}} \right\rangle ^{ \otimes 3}}\left| 1 \right\rangle \left| 0 \right\rangle \\
{\left| \psi  \right\rangle _{{{\rm{S}}_{\rm{2}}}}} = \left| {{\rm{010}}} \right\rangle {\left| {\rm{0}} \right\rangle ^{ \otimes 3}}\left| 1 \right\rangle \left| 0 \right\rangle \\
{\left| \psi  \right\rangle _{{{\rm{S}}_{\rm{3}}}}} = \left| {{\rm{011}}} \right\rangle {\left| {\rm{0}} \right\rangle ^{ \otimes 3}}\left| 1 \right\rangle \left| 0 \right\rangle\\
{\left| \psi  \right\rangle _{{{\rm{S}}_{\rm{4}}}}} = \left| {{\rm{100}}} \right\rangle {\left| {\rm{0}} \right\rangle ^{ \otimes 3}}\left| 1 \right\rangle \left| 0 \right\rangle \\
{\left| \psi  \right\rangle _{{{\rm{S}}_{\rm{5}}}}} = \left| {{\rm{100}}} \right\rangle {\left| {\rm{0}} \right\rangle ^{ \otimes 3}}\left| 1 \right\rangle \left| 0 \right\rangle
\end{array}} .\right.
\end{equation}
Next, we take ${\left| \psi  \right\rangle _{{{\rm{S}}_{\rm{0}}}}}$ as an example, then the ${H^{ \otimes 3}}$ operation is applied on the third and fourth qubits,
\begin{equation}
\label{eqn:19}
\begin{array}{l}
\left| {{\rm{000}}} \right\rangle {\left| {\rm{0}} \right\rangle ^{ \otimes 3}}\left| 1 \right\rangle \left| 0 \right\rangle \xrightarrow{H^{ \otimes 3}}{1 \over 2}\left| {{\rm{000}}} \right\rangle \sum\limits_{i = 0}^3 {\left| i \right\rangle \left| 1 \right\rangle \left| 0 \right\rangle }
\end{array}.
\end{equation}
Then we perform ${R_y}$ rotation (see Eq.~(\ref{eq6})) on the last qubit, and can get
\begin{equation}
\label{eqn:21}
{\left| \phi \right\rangle _{{{\rm{S}}_{\rm{0}}}}}={1 \over 2}\left| {{\rm{000}}} \right\rangle \sum\limits_{i = 0}^3 {\left| i \right\rangle \left| 1 \right\rangle \left( {\sqrt {{\rm{1 - }}{{\left| {{\bar v_{0i}}} \right|}^2}} \left| 0 \right\rangle  + {\bar v_{0i}}\left| 1 \right\rangle } \right)} .
\end{equation}
The other quantum states are prepared in the same way and they are listed as below,
\begin{equation}
\label{eqn:22}
\left\{ {\begin{array}{*{20}{c}}
{\left| \phi  \right\rangle _{{S_0}}} = {1 \over 2}\left| {000} \right\rangle \sum\limits_{i = 0}^3 {\left| i \right\rangle } \left| 1 \right\rangle \left( {\sqrt {1 - {{\left| {{\bar v_{0i}}} \right|}^2}} \left| 0 \right\rangle  + {\bar v_{0i}}\left| 1 \right\rangle } \right)\\
{\left| \phi  \right\rangle _{{S_1}}} = {1 \over 2}\left| {001} \right\rangle \sum\limits_{i = 0}^3 {\left| i \right\rangle } \left| 1 \right\rangle \left( {\sqrt {1 - {{\left| {{\bar v_{1i}}} \right|}^2}} \left| 0 \right\rangle  + {\bar v_{1i}}\left| 1 \right\rangle } \right)\\
{\left| \phi  \right\rangle _{{S_2}}} = {1 \over 2}\left| {010} \right\rangle \sum\limits_{i = 0}^3 {\left| i \right\rangle } \left| 1 \right\rangle \left( {\sqrt {1 - {{\left| {{\bar v_{2i}}} \right|}^2}} \left| 0 \right\rangle  + {\bar v_{2i}}\left| 1 \right\rangle } \right)\\
{\left| \phi  \right\rangle _{{S_3}}} = {1 \over 2}\left| {011} \right\rangle \sum\limits_{i = 0}^3 {\left| i \right\rangle } \left| 1 \right\rangle \left( {\sqrt {1 - {{\left| {{\bar v_{3i}}} \right|}^2}} \left| 0 \right\rangle  + {\bar v_{3i}}\left| 1 \right\rangle } \right)\\
{\left| \phi  \right\rangle _{{S_4}}} = {1 \over 2}\left| {100} \right\rangle \sum\limits_{i = 0}^3 {\left| i \right\rangle } \left| 1 \right\rangle \left( {\sqrt {1 - {{\left| {{\bar v_{4i}}} \right|}^2}} \left| 0 \right\rangle  + {\bar v_{4i}}\left| 1 \right\rangle } \right)\\
{\left| \phi  \right\rangle _{{S_5}}} = {1 \over 2}\left| {101} \right\rangle \sum\limits_{i = 0}^3 {\left| i \right\rangle } \left| 1 \right\rangle \left( {\sqrt {1 - {{\left| {{\bar v_{5i}}} \right|}^2}} \left| 0 \right\rangle  + {\bar v_{5i}}\left| 1 \right\rangle } \right)
\end{array}} .\right.
\end{equation}

Second, we randomly select a sample (assume ${\left| \phi  \right\rangle _{{{\rm{S}}_{\rm{0}}}}}$ is $\bar u$), and perform similarity calculation with other samples (i.e., ${\left| \phi  \right\rangle _{{{\rm{S}}_{\rm{1}}}}}$, ${\left| \phi  \right\rangle _{{{\rm{S}}_{\rm{2}}}}}$, ${\left| \phi  \right\rangle _{{{\rm{S}}_{\rm{3}}}}}$, ${\left| \phi  \right\rangle _{{{\rm{S}}_{\rm{4}}}}}$, ${\left| \phi  \right\rangle _{{{\rm{S}}_{\rm{5}}}}}$). Next, we take ${\left| \phi  \right\rangle _{{{\rm{S}}_{\rm{0}}}}}$ and ${\left| \phi  \right\rangle _{{{\rm{S}}_{\rm{1}}}}}$ as an example, and perform a \emph{swap} operation between the last two qubits of ${\left| \phi  \right\rangle _{{{\rm{S}}_{\rm{0}}}}}$,
\begin{equation}
\label{eqn:23}
\begin{array}{l}
{\left| \phi  \right\rangle _{{{\rm{S}}_{\rm{0}}}}}\xrightarrow{swap}\left| {{\varphi}} \right\rangle = {1 \over 2}\left| {{\rm{000}}} \right\rangle \sum\limits_{i = 0}^3 {\left| i \right\rangle  \left( {\sqrt {{\rm{1 - }}{{\left| {{\bar v_{0i}}} \right|}^2}} \left| 0 \right\rangle  + {\bar v_{0i}}\left| 1 \right\rangle } \right)\left| 1 \right\rangle}
\end{array}.
\end{equation}
After that, the \emph{swap test} operation is applied on ($\left| {{\varphi}} \right\rangle $, $\left| {{\phi}} \right\rangle_{{{\rm{S}}_{\rm{1}}}} $),
\begin{equation}
\label{eqn:24}
\begin{array}{l}
{1 \over 2}\left| 0 \right\rangle \left( {\left| \varphi  \right\rangle {{\left| {{\phi}} \right\rangle_{{{\rm{S}}_{\rm{1}}}} }} +  {{\left| {{\phi}} \right\rangle_{{{\rm{S}}_{\rm{1}}}} }}\left| \varphi  \right\rangle} \right) + {1 \over 2}\left| 1 \right\rangle \left( {{\left| \varphi  \right\rangle{\left| {{\phi}} \right\rangle_{{{\rm{S}}_{\rm{1}}}} }}  - {{\left| {{\phi}} \right\rangle_{{{\rm{S}}_{\rm{1}}}} }}\left| \varphi  \right\rangle } \right)\\
\end{array}.
\end{equation}

We perform a \emph{swap test} operation to obtain a quantum state that encodes similarity in amplitude£¬
\begin{equation}
\label{eqn:25}
\left\{ {\begin{array}{*{20}{c}}
{\left| \alpha  \right\rangle _A} = {1 \over {\sqrt 2 }}\sum\limits_{q = 1}^2 {\left| q \right\rangle } \left( {\sqrt {1 - s(\bar u,{\bar v_q})} \left| 0 \right\rangle  + \sqrt {s(\bar u,{\bar v_q})} \left| 1 \right\rangle } \right)\\
{\left| \alpha  \right\rangle _B} = {1 \over {\sqrt 2 }}\sum\limits_{q = 1}^2 {\left| q \right\rangle } \left( {\sqrt {1 - s(\bar u,{\bar v_q})} \left| 0 \right\rangle  + \sqrt {s(\bar u,{\bar v_q})} \left| 1 \right\rangle } \right)\\
{\left| \alpha  \right\rangle _C} = {1 \over {\sqrt 2 }}\sum\limits_{q = 1}^2 {\left| q \right\rangle } \left( {\sqrt {1 - s(\bar u,{\bar v_q})} \left| 0 \right\rangle  + \sqrt {s(\bar u,{\bar v_q})} \left| 1 \right\rangle } \right)
\end{array}} .\right.
\end{equation}
Then through the amplitude estimation, we can obtain the quantum states,
\begin{equation}
\label{eqn:26}
\left\{ {\begin{array}{*{20}{c}}
{\left| \beta  \right\rangle _A} = {1 \over {\sqrt 2 }}\sum\limits_{q = 1}^2 {\left| q \right\rangle } \left| {{\bar v_q} - \bar u} \right\rangle \\
{\left| \beta  \right\rangle _B} = {1 \over {\sqrt 2 }}\sum\limits_{q = 1}^2 {\left| q \right\rangle } \left| {{\bar v_q} - \bar u} \right\rangle \\
{\left| \beta  \right\rangle _C} = {1 \over {\sqrt 2 }}\sum\limits_{q = 1}^2 {\left| q \right\rangle } \left| {{\bar v_q} - \bar u} \right\rangle
\end{array}} .\right.
\end{equation}

Next, we perform an oracle operation on the quantum states obtained in the above steps to obtain the $k$ nearest neighbor samples.

\section{Simulation Experiment}

\begin{figure}[htbp]
  \centering
\centerline{\includegraphics[height=2.5cm,width=8cm,]{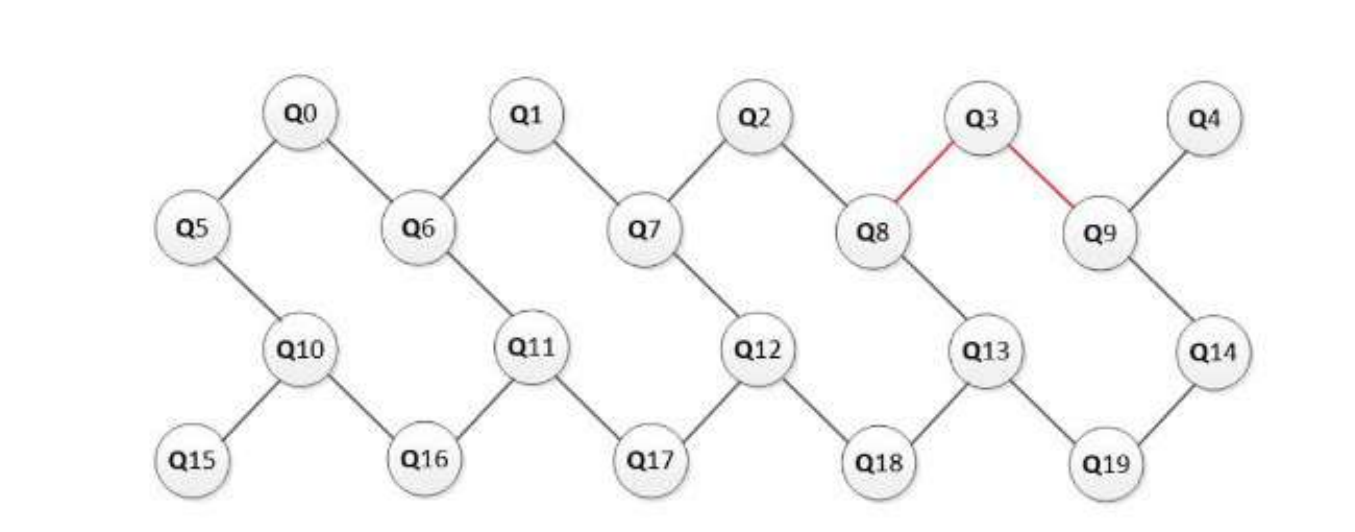}}
  \caption{The coupling map picture: Rigetti's 19-qubit processor `Acorn'. Lines indicate the two-qubit connection ruled by a controlled-$Z$ operation.}
  \label{fig07}
\end{figure}

Quantum Cloud Services ($QC{S^{TM}}$) is Rigetti's quantum-first cloud computing platform. At the end of 2017, a 19-qubit processor named `Acorn' was launched, which can be used in QCS through a quantum programming toolkit named Forest~\cite{Smith2017A}. The chip of `Acorn' is made of 20 superconducting qubits but for some technical reasons, qubit 3 is off-line and cannot interact with its neighbors, so it is treated as a 19-qubit device whose
coupling map is shown in Fig.~\ref{fig07}.

\begin{figure*}[htbp]
\centering
\includegraphics[height=10cm]{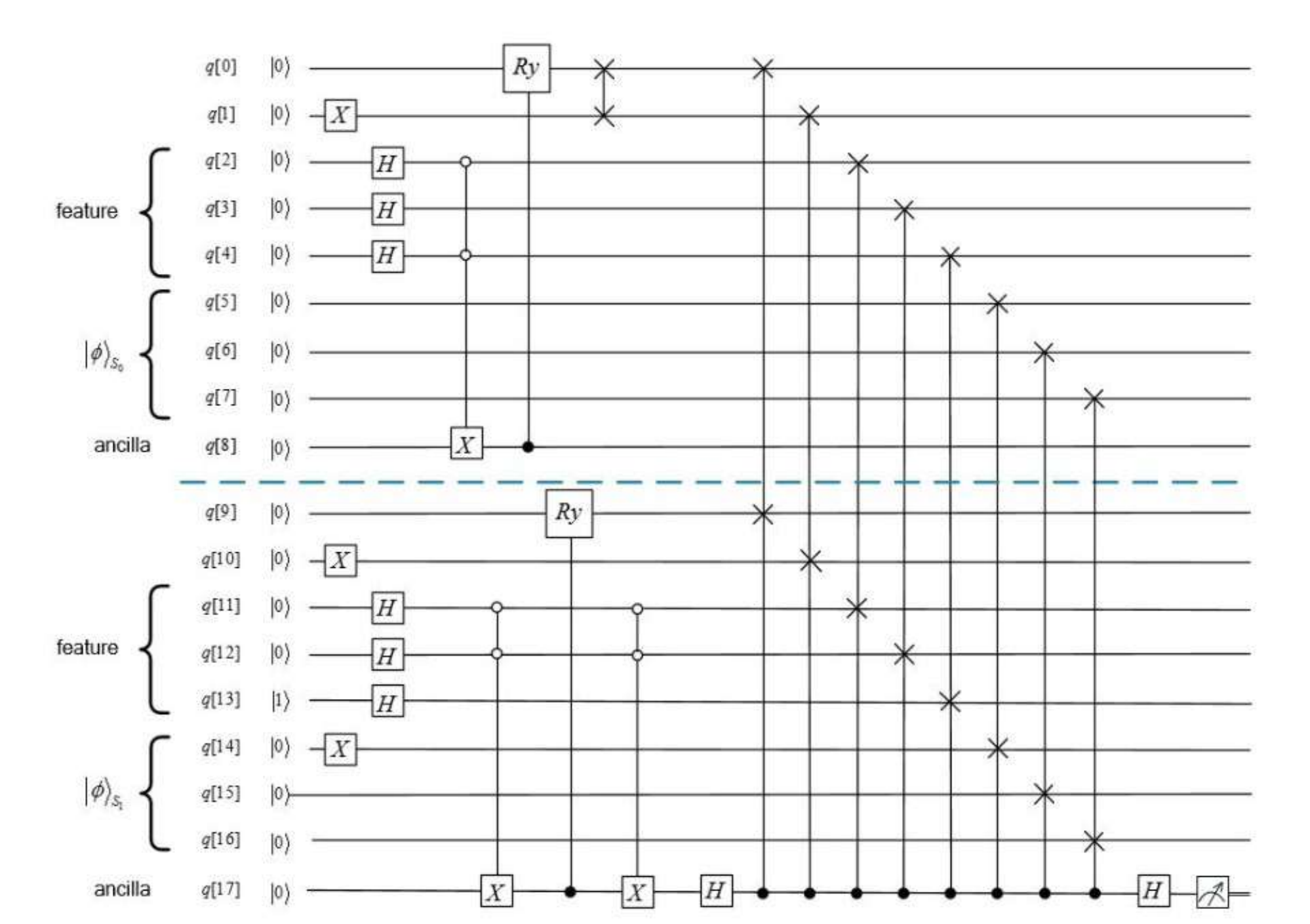}
\caption{One of the ideal experiment circuits of similarity calculation in QReliefF algorithm running on Rigetti platform. $q[0]-q[7]$ represents the randomly selected quantum state $\left| {{\phi}} \right\rangle_{S_0}$, $q[9]-q[16]$ represents $\left| {{\phi}} \right\rangle_{S_1} $, and $q[17]$ is the result qubit. \emph{X} is the Not operation, and \emph{Ry} is ${R_y}$  operation which can be expressed as a matrix
in Eq. (\ref{eq6}).}
\label{fig:5}
\end{figure*}

\begin{figure}
\centering
\includegraphics[width=3.5in]{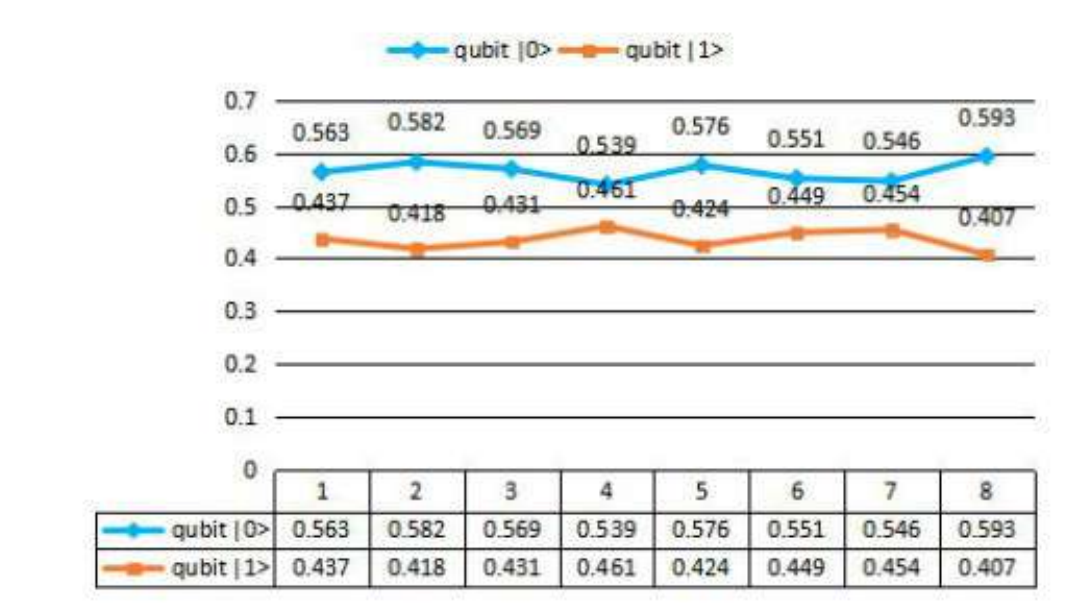}
\caption{The measurement result of $\left| 0 \right\rangle $ and $\left| 1 \right\rangle $ after running 8 times of Program 3 on Rigetti.}
\label{fig:06}
\end{figure}

In order to obtain the result and also verify our algorithm, we choose Rigetti to perform the quantum processing. However, since the Rigetti platform limits the length of the entire circuit and noise has a great influence on the preparation of quantum states~\cite{Qu2017Effectof}, we only show one of the ideal experiment circuits of similarity calculation in QReliefF algorithm running on Rigetti platform. We successfully stored the characteristic information in the sample into the amplitude of the quantum state, and then extracted the amplitude information into the quantum state through the phase estimation algorithm. Fig. \ref{fig:5} gives the schematic diagram of our experimental circuit.
The corresponding code of the circuit in Riggeti is shown in Program~\ref{SC}. After running $8$ times of Program~\ref{SC}, the result can be seen in Fig.~\ref{fig:06}. We can get $\left| 1 \right\rangle $ with the average probability of 0.435125. Then we successfully stored the characteristic information in the sample into the amplitude of the quantum state. According to Eq.~(\ref{eqn:25}), we can get $\sqrt {s\left( {\bar u,{\bar v_q}} \right)}  \approx \sqrt {0.435} $, i.e., $s\left( {\bar u,{\bar v_q}} \right) \approx 0.435$, and then extracted the amplitude information into the quantum state through the phase estimation algorithm.

\SetAlFnt{\footnotesize \sf}
\renewcommand{\algorithmcfname}{Program}
\begin{algorithm}
\SetAlgoNoLine
\caption{Similarity calculation in QReliefF algorithm running on Rigetti}
\label{SC}
$\;\;\;\;\;\;\;$\# Define the new gate from a matrix

$\;\;\;\;\;\;\;theta = Parameter('theta')$

$\;\;\;\;\;\;\;cry = np.array([$

$\;\;\;\;\;\;\;\;\;\;\;\;\;\;\;\;[1, 0, 0, 0],$

$\;\;\;\;\;\;\;\;\;\;\;\;\;\;\;\;[0, 1, 0, 0],$

$\;\;\;\;\;\;\;\;\;\;\;\;\;\;\;\;[0, 0, quil\_sqrt(1 - theta*theta), -theta*theta],$

$\;\;\;\;\;\;\;\;\;\;\;\;\;\;\;\;[0, 0, theta*theta, quil\_sqrt(1 - theta*theta)]$

$\;\;\;\;\;\;\;])$\;

$\;\;\;\;\;\;\;gate\_definition = DefGate('CRY', cry, [theta])$

$\;\;\;\;\;\;\;CRY = gate\_definition.get\_constructor()$

$\;\;\;\;\;\;\;$\# Create our program and use the new parametric gate

$\;\;\;\;\;\;\;p = Program($

$\;\;\;\;\;\;\;\;\;\;gate\_definition, X(1), H(2), H(4), H(5), X(2), X(5),$

$\;\;\;\;\;\;\;\;\;\;CCNOT(2, 5, 18), X(2), X(5), CRY(1)(18, 0), $

$\;\;\;\;\;\;\;\;\;\;SWAP(0, 1), X(10), H(11), H(12), H(13), X(14), $

$\;\;\;\;\;\;\;\;\;\;X(11), X(12), CCNOT(11, 12, 17), X(11), X(12), $

$\;\;\;\;\;\;\;\;\;\;CRY(1)(17, 9), H(19), CSWAP(19, 0, 9), $

$\;\;\;\;\;\;\;\;\;\;CSWAP(19, 1, 10), CSWAP(19, 2, 11), $

$\;\;\;\;\;\;\;\;\;\;CSWAP(19, 4, 12), CSWAP(19, 5, 13), $

$\;\;\;\;\;\;\;\;\;\;CSWAP(19, 6, 14), CSWAP(19, 7, 15), $

$\;\;\;\;\;\;\;\;\;\;CSWAP(19, 8, 16), H(19)$

$\;\;\;\;\;\;\;)$

$\;\;\;\;\;\;\;$\#print the circuit

$\;\;\;\;\;\;\;print(p)$

$\;\;\;\;\;\;\;$\# get a QPU, 20q-Acorn is just a string naming the device

$\;\;\;\;\;\;\;qc = get\_qc('20q-Acorn')$

$\;\;\;\;\;\;\;$\# run and measure

$\;\;\;\;\;\;\;result = qc.run\_and\_measure(p, trials=1024)$
\end{algorithm}

\begin{figure}
\centering
\includegraphics[width=2.2in]{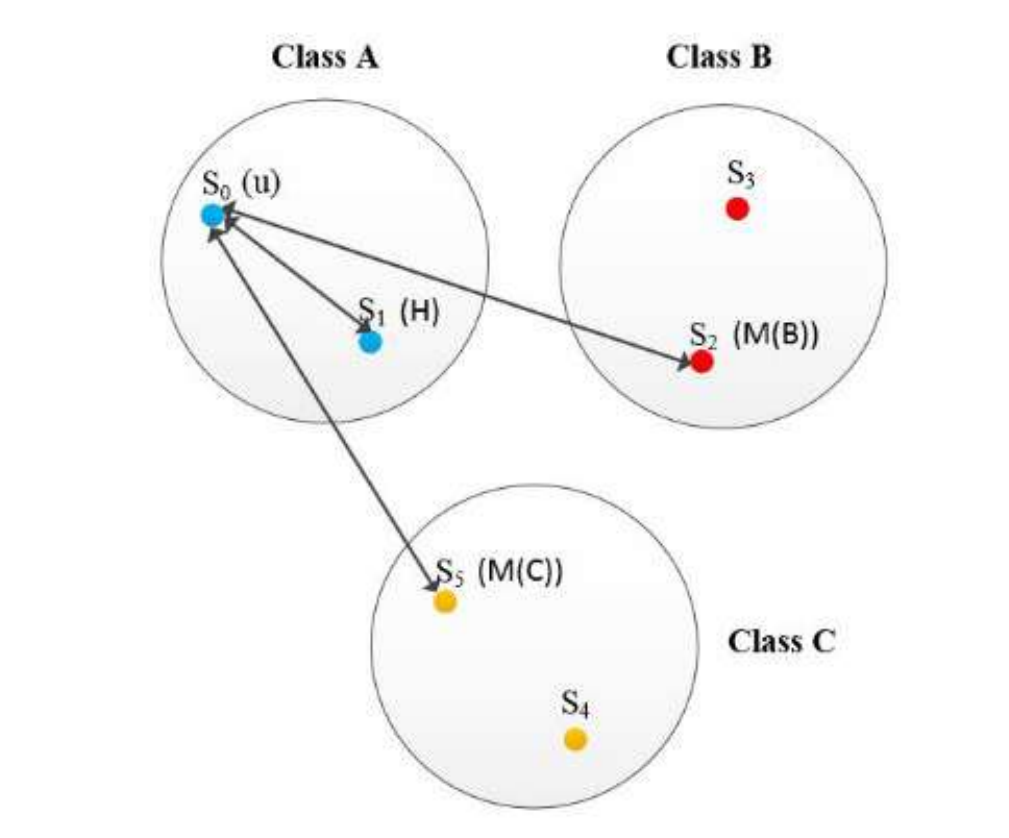}
\caption{Finding the nearest neighbor samples (${S_1}$, ${S_2}$ and ${S_5}$) of the sample ${S_0}$.}
\label{fig:6}
\end{figure}

After all the steps have been performed, we obtain the quantum states ${S_1}\;(H)$, ${S_2}\;(M(B))$ and ${S_5}\;(M(C))$ of the nearest neighbor samples of the quantum state ${S_0}\;(\bar u)$ in each class of the random sample which can be seen in Fig.~\ref{fig:6}. Then, the weight vectors are updated according to Eq.~(\ref{eq14}) and the result of $WT$ is listed in the second row of Table~\ref{tab:4} after the first iteration.
\begin{table}
\centering
\caption{The updated result of $WT$}
\label{tab:4}
\begin{tabular}{cc}
\hline\noalign{\smallskip}
$Iteration times(T)$ & $Weight vector(WT)$  \\
\noalign{\smallskip}\hline\noalign{\smallskip}
1 & [1  1 1 0 0 -1]    \\
2 & [2  2 2 -1 0 -1]   \\
3 & [3  3 3 -1 0 -2]   \\
4 & [4  4 4 -2 0 -2]   \\
\noalign{\smallskip}\hline
\end{tabular}
\end{table}
The algorithm iterates $T$ times (in our example, $T$=4) as above steps, and obtains all the $WT$ results shown in Table~\ref{tab:4}. After $T$-th iterations, $WT=[4,4,4,-2,0,-2]$, then $\overline{WT}=[1,1,1,-1/2,0,-1/2]$. In this paper, the value of $\tau$ in the example is assumed to be 0.5 according to the updated result of WT in TABLE~\ref{tab:4}. Since the threshold $\tau=0.5 $, so the selected features are $F_0$, $F_1$ and $F_2$, i.e., the first, second and third features. The result of quantum feature selection is consistent with the classical ReliefF algorithm after being verified by python.

In the final weight value comparison, considering the large amount of data in the complex system and the corresponding eigenvalues, the calculation amount required for the comparison after the final result is obtained is also large. In order to meet the requirements of big data and result accuracy, we adopted an optimized quantum maximum and minimum value search algorithm~\cite{Chen2019An} when comparing weights in the last step to help us quickly and accurately select the features we want, so as to better solve the multi-classification problem in complex systems.

In circumstances when we can exactly estimate the ratio of the number of solutions $M$ and the searched space $N$, this algorithm can improve the successful probability close to $100{\rm{\% }}$. Furthermore, it shows an advantage in complexity with large databases and in the operation complexity of constructing oracles.

\section{Efficiency analysis}

In order to evaluate the efficiency of QReliefF algorithm, three algorithms (i.e., classical Relief, classical ReliefF, quantum Relief algorithms) are selected to compare with our algorithm from three indicators: complexity of similarity calculation (CSC), complexity of finding the nearest neighbor (CFNN) and resource consumption (RC).

In the classic Relief algorithm, it takes $O(N)$ time to calculate the distance between randomly selected samples and any other samples, and then finds the nearest neighbors related to $M$. This process needs to iterate $T$ times, so CSC is $O(TMN)$. Since $T$ is a constant, then CSC in the classic Relief algorithm is $O(MN)$. As we know, there are totally $M$ samples, each with $N$ features, so CFNN is $O(M)$ and RC in the classic Relief algorithm is $O(MN)$ bits. The classical ReliefF algorithm is similar to the classical Relief algorithm. Since it finds $k$ nearest neighbors at once time, so the time complexity is $O(kTMN)$. Then we can simplify CSC to $O(MN)$ because $k$ and $T$ are constants. Besides, CFNN for finding $k$ nearest neighbors is $O(M)$. In terms of resource consumption, there are $M$ samples, each sample has $N$ features, so the resource consumption of the classic ReliefF algorithm is $O(MN)$ bits.

In QRelief and QReliefF algorithms, the quantum property is used to calculate the distance from $O(N)$ to $O(1)$, so their CSCs are all $O(TM)$. Since $T$ is constant, their CSCs can be simplified to $O(M)$. CFNN of QRelief is $O\left( {kM } \right)$, then it can be simplified to $O\left( {M } \right)$ as $k$ is constant. While CFNN of QReliefF is $\sqrt {kM} $ because it uses the quantum Grover-Long method to find $k$ nearest neighbor samples which holds a quadratic acceleration. Since $k$ is constant, CFNN of QReliefF is $O(\sqrt M )$. RC of similarity calculation, finding the nearest neighbor samples and updating weight vector are $O(TMlogN)$, $O(TN)$ and $O(N)$, respectively. Therefore, the total complexity is $O(TMlogN+TN+N)$, since $T$ is constant, then RC of QRelief and QReliefF are all $O(MlogN+N)$. For multi-features big data in complex systems with edge computing, there is $M \gg N$, so $M \gg {N \over {N - \log N}}$, then RC of QRelief and QReliefF can be simplified into $O(MlogN)$.

For convenience, we list the efficiency comparison of classic Relief algorithm, ReliefF algorithm, quantum Relief algorithm and our algorithm in terms of CSC, CFNN and RC in Table~\ref{tab1}. Obviously, our algorithm is superior than classical algorithms (i.e., Relief and ReliefF) in terms of CSC, CFNN and RC, and better than quantum algorithm (i.e., QRelief) in terms of CFNN.

\begin{table}[htbp]
\caption{Efficiency comparison between classical Relief, classical ReliefF, QRelief and our QReliefF algorithms}
\centering
\begin{tabular}{ccccc}
\hline
& Multi-classification & CSC & CFNN & RC \\
\hline
Relief  & no & $O(MN)$  & $O(M)$  & $O(MN)$ \\
ReliefF & yes & $O(MN)$  & $O(M)$  & $O(MN)$ \\
QRelief & no & $O(M)$   & $O(M)$  & $O(MlogN)$\\
QReliefF & yes & $O(M)$  & $O(\sqrt M )$  & $O(MlogN)$\\
\hline
\end{tabular}
\label{tab1}
\end{table}

\section{Conclusion and discussion}

With the rapid development of edge computing technology and quantum machine learning algorithms, researchers began to pay attention to the combination and application of these two fields. In this paper, we use quantum technology to solve the multi-classification problem of feature selection in the complex systems with edge computing, and propose a quantum ReliefF algorithm. Compared to the classic ReliefF algorithm, our algorithm reduces the complexity of similarity calculation from $O(MN)$ to $O(M)$, and the complexity of finding the nearest neighbor from $O(M)$ to $O(\sqrt M )$. In addition, from the perspective of resource consumption, our algorithm consumes $O(MlogN)$ qubit, while the classic ReliefF algorithm consumes $O(MN)$ bit. Obviously, our algorithm is superior in terms of computational complexity and resource consumption.

It should be noted that our work aims to improve the algorithm efficiency, while the privacy protection of sensitive data is not taken into account. At present, data security has become a focus of attention in the field of artificial intelligence, some solutions for data privacy protection in complex systems with edge computing have been proposed~\cite{Xu2019Joint}\cite{Xiong2019Partially}\cite{Zhang2018Data}\cite{Xu2019Anedge}. In our future work, how to improve the efficiency of quantum machine learning algorithms while ensuring the privacy protection of sensitive data, such as~\cite{Liu2018FullBlind}\cite{Liu2019QuantumSearchable}\cite{Liu2019Privacy}\cite{Liu2019ANovel}, will become our direction.



\section*{Acknowledgment}
The authors would like to express heartfelt gratitude to the anonymous reviewers and editor for their comments that improved the quality of this paper. And the support of all the members of the quantum research group of NUIST \& SEU is especially acknowledged, their professional discussions and advices have helped us a lot.

\vspace{12pt}

\end{document}